\documentclass[runningheads]{llncs}
\usepackage{subcaption}
\usepackage{graphicx}
\usepackage{algorithm} 
\usepackage{algpseudocode}
\usepackage{lineno,hyperref}
\usepackage{dblfloatfix}
\usepackage{amsmath,amssymb,amsfonts}
\usepackage{booktabs}
\usepackage{float}
\usepackage{array}
\newcolumntype{?}{!{\vrule width 1pt}}

\begin{document}

\title{ViMQ: A Vietnamese Medical Question Dataset for Healthcare Dialogue System Development}
\author{Ta Duc Huy \inst{1} \and
Nguyen Anh Tu\inst{1} \and
Tran Hoang Vu\inst{1} \and
Nguyen Phuc Minh\inst{1} \and
Nguyen Phan\inst{1} \and
Trung H. Bui \and %\email{trungbui@gmail.com}
Steven Q. H. Truong\inst{1}
}

\authorrunning{Huy et al.}
\institute{VinBrain \\
\email{\{v.huyta, v.tunguyen, v.vutran, v.minhng, v.nguyenphan, v.brain01\}@vinbrain.net} \\
\email{bhtrung@gmail.com} \\
%\url{http://www.springer.com/gp/computer-science/lncs}
}
%utran
\titlerunning{ViMQ: Vietnamese Medical Question dataset for healthcare }
\maketitle
\begin{abstract}
Existing medical text datasets usually take the form of question and answer pairs that support the task of natural language generation, but lacking the composite annotations of the medical terms. 
In this study, we publish a Vietnamese dataset of medical questions from patients with sentence-level and entity-level annotations for the Intent Classification and Named Entity Recognition tasks.
The tag sets for two tasks are in medical domain and can facilitate the development of task-oriented healthcare chatbots with better comprehension of queries from patients.
We train baseline models for the two tasks and propose a simple self-supervised training strategy with span-noise modelling that substantially improves the performance.
Dataset and code will be published at {\url{https://github.com/tadeephuy/ViMQ}}.
\end{abstract}
\section{Introduction}
Named Entity Recognition (NER) involves extracting certain entities in a sentence and classifying them into a defined set of tags such as organizations, locations, dates, time, or person names. In the medical domain, NER tag sets usually contain patient information and clinical terms. Intent Classification (IC) is the task of categorizing a sequence into a set of defined intentions. In the medical domain, IC classes could include the conversation objectives in interactions between patients and clinical experts. NER and IC are two major components of a task-oriented dialogue system. There exist several medical conversation datasets in English \cite{he2020meddialog}, German \cite{rojowiec-etal-2020-intent}, and Chinese \cite{he2020meddialog,zeng-etal-2020-meddialog}, while Vietnamese medical conversation datasets are not abundant.
COVID-19 NER for Vietnamese \cite{PhoNER_COVID19} is a dataset for the named entity recognition task with generalized entity types that can extend its application to other future epidemics. The dataset is crawled from Vietnamese online news websites with the keyword “COVID” and filtered out irrelevant sentences, resulting in 34,984 entities and 10,027 sentences. 
% The entity classes in the dataset include PATIENT\_ID, PERSON\_NAME, AGE, GENDER, OCCUPATION, LOCATION, ORGANIZATION, SYMPTOM\&DISEASE, TRANSPORTATION and DATE. 
% It employs a golden annotated set of 1000 sentences annotated by the authors of the guidelines with a cross-validation session to reach a consensus and then use it as a probing set to validate the correctness of other subsets distributed to other subsets annotators.

To our knowledge, we are the first to publish a Vietnamese Medical Question dataset (ViMQ) that contains medical NER and IC tags set where the applications could be generalized to developing the Natural Language Understanding (NLU) module for healthcare chatbots.

NER annotation includes highlighting the span of a given entity and assigning an appropriate tag. Specifying the spans of the entities can raise consensus issues stemming from the subjectivity of different annotators while classifying entities into tags are far more compliant due to the distinctiveness of the tags. In this paper, we developed an annotation methodology that aims to minimize this effect. In addition, NER annotation tools with poor ergonomic design can contribute to the span-noise, where the start and end indexes of a given span are shifted from their correct ground truth. We propose a training strategy to learn the model with such noise. To summarize, our contributions are:

\begin{itemize}
    \item  We published a Vietnamese Medical Question dataset for healthcare chatbot development.
    \item We proposed a training strategy to learn the model with span-noise for the NER task.
\end{itemize}

\section{ViMQ dataset}

\subsection{Intent and entity types}
The ViMQ dataset contains Vietnamese medical questions crawled from the consultation section online between patients and doctors from \url{www.vinmec.com}, a website of a Vietnamese general hospital. Each consultation consists of a question regarding a specific health issue of a patient  and a detailed respond provided by a clinical expert. The dataset contains health issues that fall into a wide range of categories including common illness, cardiology, hematology, cancer, pediatrics, etc. We removed sections where users ask about information of the hospital and selected 9,000 questions for the dataset. We annotated the questions for the NER and IC tasks. The tag sets for two tasks could be applied to a dialogue system for medical consultation scenario, where the chatbot acts as a medical assistant that inquires queries from users for their health issues. Each question is annotated for both tasks. The statistics of the labels are shown in Table \ref{table: vimq-stats}. 
\paragraph{Entity definition:} The tag set for NER includes SYMPTOM\&DISEASE, MEDICAL\_PROCEDURES and MEDICINE:
\begin{itemize}
    \item SYMPTOM\&DISEASE: any symptom or disease that appears in the sentence, including disease that are mentioned as part of a medicine such as “rabies” in “rabies vaccine”.
    \item MEDICAL\_PROCEDURE: actions provided by clinical experts to address the health issue of the patient, including diagnosing, measuring, therapeutic or surgical procedures.
    \item MEDICINE: any medicine name that appears in the sentence.
\end{itemize}

\begin{table}[!t]
\centering
\caption{Statistics of ViMQ dataset.} 
\label{table: vimq-stats}
\begin{tabular}{l|r|r|r|r}
\hline
\textbf{Entity Type}           & \textbf{Train} & \textbf{Valid} & \textbf{Test} & \textbf{All} \\ \hline
SYMPTOM\&DISEASE          & 10,599          & 1,300           & 1,354          & 13,253        \\ 
MEDICAL\_PROCEDURE             & 1,583           & 204            & 213           & 2,000         \\ 
MEDICINE                           & 781            & 90             & 108           & 979          \\ \hline

\textbf{Intent Type}           & \textbf{Train} & \textbf{Valid} & \textbf{Test} & \textbf{All} \\ \hline
Diagnosis  & 3,444 & 498 & 484 & 4,426 \\
Severity & 1,070 & 150 &  162 & 1,382 \\ 
Treatment & 1,998 & 264 & 265 & 2,527 \\
Cause & 488 & 88 & 89 & 625 \\\hline

% \# Entities in total  & 12,963          & 1,594           & 1,675          & 16,232        \\ \hline
% \# Sentences in total & 7,000           & 1,000           & 1,000          & 9,000        \\  \hline
% \# Avg tokens of sentence & 12 & 12 & 12 & 12                                              \\  \hline

\end{tabular}
\end{table}

\paragraph{Intent definition:} The tag sets for IC includes Diagnosis, Severity, Treatment and Cause:
\begin{itemize}
    \item Diagnosis: questions relating to identification of symptoms or diseases.
    \item Severity: questions relating to the conditions or grade of an illness. 
    \item Treatment: questions relating to the medical procedures for an illness. 
    \item Cause: questions relating to factors of a symptom or disease. 
\end{itemize}

\begin{figure}[hbt!]
    \centering
    \includegraphics[width=0.8\linewidth]{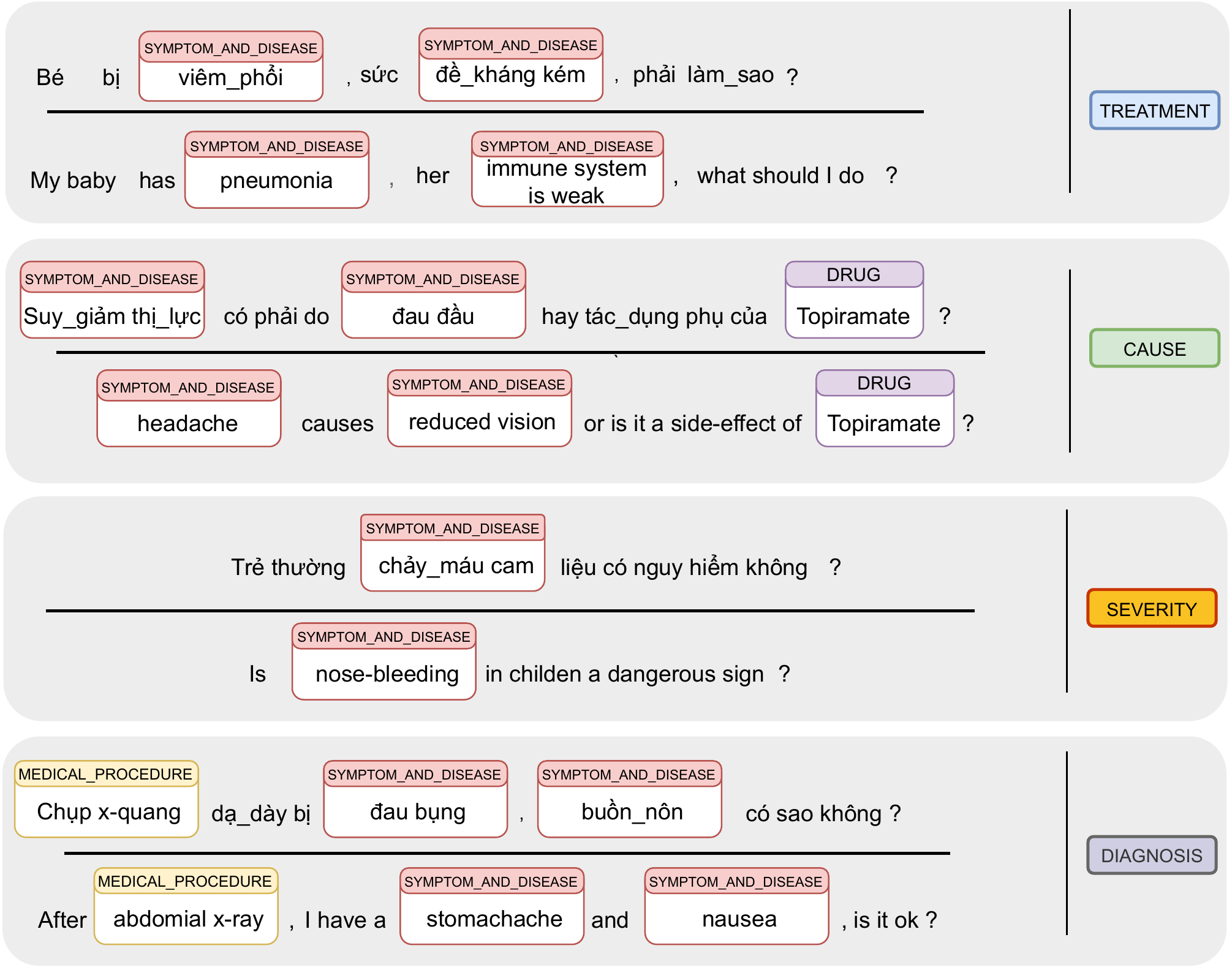}
    \caption{Examples from ViMQ dataset. Each block includes the IC tag (right side) and the NER tags of an example from the dataset (above the horizontal line) and its English translation (below the horizontal line).}
    \label{fig: vimq-example}
\end{figure}

\subsection{Annotation process}
We propose a novel method to manage the annotation process for the two tasks of NER and IC called Hierarchical Supervisors Seeding (HSS). The method circumvents around building a solid and consensual supervisor team which grows in size as the annotation process progresses. Leveraging a solid supervisor team mitigates the subjectivity and improves the task-oriented competency of individual annotators. Each supervisor, upon designated, does not only make mutual concessions for label disagreements of individual annotators but also acts as a seed to make way for them to become the next generation supervisors. 

We first ran a pilot phase where 1,000 samples are selected for annotation, called the pilot set. The author of the annotation guideline is designated as the first supervisor, where two other NLP engineers are individual annotators. The two individual annotators follow the annotation guideline and annotate two overlapping sets of 600 sentences in the pilot set. The intersection of the two sets consisting of 100 sentences is also annotated by the supervisor and is used to measure the annotation quality of the two NLP engineers using F1-score. The intersection set is selected such that it covers a broad range of difficulties. Hard sentences contain rare diseases such as anxiety disorders, thalassemia. Easy sentences contain common diseases or symptoms such as flu, fever, and cold. Both individual annotators are required to achieve an F1-score of at least 0.9 on both tasks on the intersection set before discussing with the supervisor on the disagreement cases. The supervisor has to point out which parts of the annotation guideline could solve these conflicts and update the guideline until they reach a consensus. This session encourages the individual annotators to gain better comprehension and help the supervisor to improve the guideline. 

Two individual annotators in the pilot phase are designated as supervisors in the following phases. In the main phase, each supervisor monitors two other individual annotators for a set of the next 1,000 sentences. They follow the same procedure in the pilot phase where each of the individual annotators works with 600 sentences, and the conflicts in the intersection set of 100 sentences are solved by the supervisor of each group in the discussion sessions. All updates to the annotation guideline made by the supervisors are reviewed by the guideline author. We continue to designate the two individual annotators to be supervisors in the next phase in a hierarchical manner. We repeat the process until all sentences are annotated. The intersection sets in each group are aggregated and used as the golden test set. The individual annotators after the pilot phase are medical under-graduates, which are hired at a rate of 0.1 USD per annotated sentence and 0.05 USD per resolved conflict. 

\section{Baseline models} \label{baseline}
\subsection{Intent classification}
For task intent classification, we follow a common strategy when fine tuning pretrained BERT \cite{devlin-etal-2019-bert} for sequence classification task, we use PhoBERT \cite{phobert} to extract contextual features of sentences then take the hidden state of the first special token ([CLS]) denoted $h_1$, the intent is predicted as:
\begin{equation}
    y^i = softmax(W^i h_1 + b^i)
\end{equation}

\subsection{Named entity recognition}
A standard approach for NER task is to formulate it as a sequence labeling problem with the Begin-Inside-Outside (BIO) tagging scheme. Similar to the intent classification task, we use PhoBERT to extract contextual embedding with a conditional random fields (CRF) \cite{10.5555/645530.655813} inference layer on top of our model to improve sequence labeling performance. 

\textbf{Sub-word Alignment}: In our approach, because PhoBERT uses a Byte Pair Encoder (BPE) to segment the input sentence with sub-word units. A word is split into k sub-words. The original word label is assigned to the first sub-word and k-1 sub-words is marked with a special label (X). Final, the predicted label of original word based on the first label of sub-word, we keep a binary vector for active positions.

\section{Method} \label{method}
As the annotators only agree on the intersection sets, the remaining sets, which were only labeled by single annotators, could be polluted with noise.
We develop a training strategy to minimize the effect of remaining noisy labels in the training set.
We empirically show that the method makes use of the potentially noisy samples and improves the performance substantially.
We apply our training strategy on the ViMQ dataset and the COVID-19 NER Vietnamese dataset and achieve better performance using the standard settings.

Given a set of entities $\{E_i\}$ in a sentence, where $E_i$ is a tuple of $(s_i, e_i, c_i)$ where $s_i$ and $e_i$ is the start and end index of the named entity $i^{th}$ and $c_i$ is its category.

\textbf{Span-noise modelling}. We model the span-noise by adding $\delta$ to $s_i$ and $e_i$ with a probability of $p$ during training. Span-noise modeling acts as a regularization method and possibly corrects the noisy span indexes during training model.

\textbf{Online self-supervised training}. The training progresses through $N$ iterations, each consisting of $T$ epochs.
After training for the first iteration, we start to aggregate the predictions made by the model in each epoch in the $j^{th}$ iteration.
Entities with correct predictions for the entity category $c_i$ and have an $IoU > 0.4$ with the span ground truth $(s_i, e_i)$ are saved.
We then employ major voting for the start/end indices of the span of each entity to combine the aggregated predictions of $T$ epochs in iteration $j^{th}$ and use the result as labels for training the model in the next iteration.

% \begin{algorithm}
% 	\caption{Online self-supervised training with noise.}
% 	\label{algo1}
% 	\begin{algorithmic}[1]
% 	    \State \textbf{Input} training\ set\ $D = \{x_i, y_i\}$, epochs\ per\ iteration\ $T$,\ noise\ probability\ p, noise\ $\delta$, model $f_\theta$
% 	    \State \textbf{Init} iteration prediction\ list\ $P_j = \{\hat{y_i}\}$\\
% 		\For {iteration $j=1,2,\ldots,N$}
% 			\State $y_i$ $\leftarrow$ Add noise $\delta$ with probability of $p$ to $y_i \in D$
% 			\For {$epoch=1,2,\ldots,T$}
% 			    \For {$x_i, y_i \in D$}
% 			        \State $\hat{y_i}$ = $f_\theta(x_i)$
%     			    %\State Get prediction $\hat{y_i}$ from model with input $x_i$
%     			    % bien dong' nay` thanh` cong thuc
%     			    \State $\theta \leftarrow \theta - \grad_\theta (- y_i \log (\hat{y_i}))$
%     			    \State $P_j \leftarrow P_j \cup \{\hat{y_i}\}$
%     			    %\State Add $\hat{y_i}$ to $P_j$
% 			    \EndFor
% 			\EndFor
% 			\If{$j > 1$}
% 			    \State $y_i \leftarrow$ major\_voting($P_j$)
% 			\EndIf
% 		\EndFor
% 	\end{algorithmic} 
% \end{algorithm}

\section{Experiments}
\subsection{Experiments setups}
We conduct experiments on our dataset and the COVID-19 NER dataset to compare the performances of the baseline models and the online self-supervised training strategy for the IC and NER tasks which are presented in Sections \ref{baseline} and \ref{method}. It should be noted that we do not add noise in the IC task because span-noise is inapplicable. Our experimental models  were implemented PyTorch \cite{pytorch} using Huggingface's Transformers \cite{wolf-etal-2020-transformers}, a huge library for pretrained Transformer-based models. We train both of the models for 5 iterations, each with 10 epochs. In the online self-supervised approach, the model starts using pseudo-labels from the second iteration. We set the noise injection probability $p$ to 0.1 and the noise shifting offset $\delta$ to 1 in all of our experiments. We employ AdamW optimizer with learning rate at $5e^{-5}$. In this study, we ran each experiments 5 times with different random seeds and report the average values with their standard deviations.
\subsection{Results}

\subsubsection{Intent classification.}
Table \ref{table:ic_result} shows that the online self-supervised method improves baseline model from  90.36\% to 90.65\% Micro-F1 and from 91.17\% to 91.65\% Macro-F1.
The confusion matrix in Table \ref{confusion_matrix} shows that the IC model makes mistakes some of the sample having intent Diagnosis of intent Treatment and vice versa.

% \begin{table}[!t]
% \centering
% \caption{F1-score of IC task on ViMQ dataset.}
% \vspace{-0.2cm}
% \label{table:ic_result}
% \begin{tabular}{c|c|c?c|c|c|c}
% \hline
% \toprule[1pt]
%     & \multicolumn{2}{c?}{F1-Score(\%)} &  & \multicolumn{3}{c}{F1-Score(\%)}\\ \hline
% \textbf{Baseline}  & \checkmark &     \checkmark  &  \textbf{Baseline} & \checkmark &     \checkmark     & \checkmark\\  \hline
% + \textbf{self-sup.}   & - &     \checkmark   & + \textbf{self-sup.}  & - &     \checkmark     & \checkmark \\ \hline

% Diagnosis & 90.22 $\pm$ 0,05             & \textbf{90.55 $\pm$ 0,09}   &   + \textbf{span-noise} & - & - & \checkmark \\ \hline
% Severity  & 90.22 $\pm$ 0,38              & \textbf{91.02 $\pm$ 0,00}      & SYMP.\&DIS.    & 73.44 $\pm$ 0.05 & 73.60 $\pm$ 0.29          & \textbf{77,21 $\pm$ 0.09} \\ \hline
% Treatment & \textbf{89.04 $\pm$ 0,23}    & 88.97 $\pm$ 0,21             & MED.    & 58.14 $\pm$ 2.47 & 63.22 $\pm$ 0.47          & \textbf{67,34 $\pm$ 0.51} \\ \hline
% Cause     & 95,15 $\pm$ 0,9            & \textbf{96.05 $\pm$ 0,00}      & MED.\_PRO.    & 58.51 $\pm$ 0.41 & 59.01 $\pm$ 0.78 & \textbf{61,73 $\pm$ 0.08}          \\ \hline
% Micro-F1  & 90.36 $\pm$ 0,05             & \textbf{90.65 $\pm$ 0,15}    & Micro-F1 & 70.78 $\pm$ 0.20  & 71.22 $\pm$ 0.30          & \textbf{74,78 $\pm$ 0.05} \\ \hline
% Macro-F1  & 91,17 $\pm$ 0,23             & \textbf{91.65 $\pm$ 0,08}     & Macro-F1 & 63.36 $\pm$ 0.70  & 65.28 $\pm$ 0.51          & \textbf{68,76 $\pm$ 0.23} \\ 
% \bottomrule[1.5pt]
% \end{tabular}
% \end{table}

\begin{table}[!t]

\begin{minipage}[t]{.5\linewidth}
\centering
\captionsetup{justification=centering}
\caption{ Confusion matrix of the\\ IC task on the ViMQ dataset.}

\begin{tabular}{l|c|c|c|c}

\hline
\toprule[1pt]
          & \multicolumn{1}{|l|}{Sev.} & \multicolumn{1}{|l|}{Cau.} & 
          \multicolumn{1}{|l|}{Tre.} &
          \multicolumn{1}{|l}{Dia.} \\ 
\midrule[1pt]
Sev.  & 152                          & 0                         & 0                             & 10                            \\ \hline
Cau.     & 0                            & 85                        & 0                             & 4                             \\ \hline
Tre. & 1                            & 1                         & 233                           & 30                            \\ \hline
Dia.  & 19                           & 2                         & 27                            & 436                           \\ \bottomrule[1.5pt]
\end{tabular}
\label{confusion_matrix}
\end{minipage}%
\hfill
\begin{minipage}[t]{.5\linewidth}
\captionsetup{justification=centering}
\caption{F1-score of IC task on ViMQ dataset.}
\centering
\label{table:ic_result}
\begin{tabular}{c|c|c}
\hline
\toprule[1pt]
    & \multicolumn{2}{c}{F1-Score(\%)}\\ \hline
\textbf{Baseline}  & \checkmark &     \checkmark  \\ \hline
+ \textbf{self-sup.}   & - &     \checkmark   \\ \hline
Dia. & 90.22 $\pm$ 0.05             & \textbf{90.55 $\pm$ 0,09}    \\ \hline
Sev.  & 90.22 $\pm$ 0.38              & \textbf{91.02 $\pm$ 0,00}      \\ \hline
Tre. & \textbf{89.04 $\pm$ 0,23}    & 88.97 $\pm$ 0.21            \\ \hline
Cau.     & 95.15 $\pm$ 0.9            & \textbf{96.05 $\pm$ 0.00}          \\ \hline
Mic.F1  & 90.36 $\pm$ 0.05             & \textbf{90.65 $\pm$ 0.15}    \\ \hline
Mac.F1  & 91.17 $\pm$ 0.23             & \textbf{91.65 $\pm$ 0.08}      \\ 
\bottomrule[1.5pt]
\end{tabular}

\end{minipage} 
\end{table}

\subsubsection{Named entity recognition}
Tables \ref{table:ner-covid} and \ref{table:ner-vimq} show the experiments results on the COVID-19 NER and ViMQ datasets for the NER task. Online self-supervised training consistently improves the F1-score compares to the baseline model. The injection of span-noise adds significant gain to the performance on both datasets with a margin of 1\% on COVID-19 NER and 3\% on ViMQ
In ViMQ dataset, there are 389 sentences with wrong predictions. Boundary errors exists in 295/389 sentences that contains 362 entities with correct categories but incorrect span indices. Most of them are from class SYMPTOM\&DISEASE as labels sometimes contain the inflicted body parts of the patient.

% \begin{table}[]
% \caption{Experimental results of NER task}
% \resizebox{\textwidth}{!}{
% \centering
% \begin{tabular}{|l|l|l|l|l|l|l|l|l|l|l|l|l|}
% \hline
%         Model & AGE. & DAT. & GEN. & JOB. & LOC. & NAM. & ORG. & PAT. & SYM. & TRA. & Micro-F1 & Macro-F1 \\ 
% \hline
%         baseline & 92,26 $\pm$ 2,02 & 98,12 $\pm$ 0,05 & 93,41 $\pm$ 1,12 & 76,47 $\pm$ 3,53 & 92,90 $\pm$ 0,36 & 91,49 $\pm$ 0,02 & 87,92 $\pm$ 1,17 & 94,94 $\pm$ 0,83 & 86,98 $\pm$ 0,63 & 97,03 $\pm$ 1,42 & 93,34 $\pm$ 0,04 & 91,15 $\pm$ 0,81 \\ 
% \hline
%         baseline + update & 91,59 $\pm$ 2,04 & 98,13 $\pm$ 0,03 & 94,81 $\pm$ 1,26 & 74,52 $\pm$ 1,66 & 94,13 $\pm$ 0,80 & 91,98 $\pm$ 0,84 & 89,21 $\pm$ 1,23 & 95,70 $\pm$ 0,84 & 88,61 $\pm$ 1,27 & 96,85 $\pm$ 0,62 & 94,34 $\pm$ 0,52 & 91,55 $\pm$ 0,81 \\ 
% \hline
%         baseline + update + noise & 94,08 $\pm$ 0,77 & 98,25 $\pm$ 0,36 & 95,06 $\pm$ 0,45 & 77,15 $\pm$ 3,81 & 94,20 $\pm$ 0,28 & 94,03 $\pm$ 0,96 & 90,22 $\pm$ 0,45 & 96,28 $\pm$ 0,84 & 87,77 $\pm$ 0,49 & 97,84 $\pm$ 0,66 & 94,81 $\pm$ 0,33 & 92,49 $\pm$ 0,22 \\ 
%     \hline
%     \end{tabular}}
% \end{table}

\begin{table}[!t]
\caption{F1-score of NER task on COVID-19 Vietnamese dataset.}
% \vspace{-0.2cm}
\label{table:ner-covid}
\centering
\begin{tabular}{c|c|c|c}
\hline
% Model    & \begin{tabular}[c]{@{}c@{}}\textbf{Baseline} \\ \textbf{(\%)}  \end{tabular}       & \begin{tabular}[c]{@{}c@{}} \textbf{Baseline} \\ \textbf{+self.sup.} \\ \textbf{(\%)}     \end{tabular}    & \begin{tabular}[c]{@{}c@{}} \textbf{Baseline}\\ \textbf{+self.sup.} \\ \textbf{+noise} \\ \textbf{(\%)} \end{tabular} \\ \hline
\toprule[1pt]
    & \multicolumn{3}{c}{F1-Score(\%)} \\ \hline
\textbf{Baseline}  & \checkmark &     \checkmark     & \checkmark\\  \hline
+ \textbf{self-supervised}   & - &     \checkmark     & \checkmark\\ \hline
+ \textbf{span-noise}   & - &       -   & \checkmark \\ 
\midrule[1pt]
% AGE.     & 92.26 $\pm$ 2.02 & 91.59 $\pm$ 2.04          & \textbf{94.08 $\pm$ 0.77} \\ \hline
% DAT.     & 98.12 $\pm$ 0.05 & 98.13 $\pm$ 0.03          & \textbf{98.25 $\pm$ 0.36} \\ \hline
% GEN.     & 93.41 $\pm$ 1.12 & 94.81 $\pm$ 1.26          & \textbf{95.06 $\pm$ 0.45} \\ \hline
% JOB.     & 76.47 $\pm$ 3.53 & 74.52 $\pm$ 1.66          & \textbf{77.15 $\pm$ 3.81} \\ \hline
% LOC.     & 92.90 $\pm$ 0.36 & 94.13 $\pm$ 0.80          & \textbf{94.20 $\pm$ 0.28} \\ \hline
% NAM.     & 91.49 $\pm$ 0.02 & 91.98 $\pm$ 0.84          & \textbf{94.03 $\pm$ 0.96} \\ \hline
% ORG.     & 87.92 $\pm$ 1.17 & 89.21 $\pm$ 1.23          & \textbf{90.22 $\pm$ 0.45} \\ \hline
% PAT.     & 94.94 $\pm$ 0.83 & 95.70 $\pm$ 0.84          & \textbf{96.28 $\pm$ 0.84} \\ \hline
% SYM.     & 86.98 $\pm$ 0.63 & \textbf{88.61 $\pm$ 1.27} & 87.77 $\pm$ 0.49          \\ \hline
% TRA.     & 97.03 $\pm$ 1.42 & 96.85 $\pm$ 0.62          & \textbf{97.84 $\pm$ 0.66} \\ \hline
Mic.F1 & 93.34 $\pm$ 0.04 & 94.34 $\pm$ 0.52          & \textbf{94.81 $\pm$ 0.33} \\ \hline
Mac.F1 & 91.15 $\pm$ 0.81 & 91.55 $\pm$ 0.81          & \textbf{92.49 $\pm$ 0.22} \\ 
\bottomrule[1.5pt]
\end{tabular}
\end{table}

\begin{table}[!t]
\centering
\caption{F1-score of the NER task on the ViMQ dataset.}
% \vspace{-0.2cm}

\label{table:ner-vimq}
\begin{tabular}{c|c|c|c}
\hline
\toprule[1pt]
    & \multicolumn{3}{c}{F1-Score(\%)} \\ \hline
\textbf{Baseline} & \checkmark & \checkmark & \checkmark \\  \hline
+ \textbf{self-supervised} & - & \checkmark & \checkmark\\ \hline
+ \textbf{span-noise} & - & - & \checkmark \\
\midrule[1pt]
SYMP.\&DIS.    & 73.44 $\pm$ 0.05 & 73.60 $\pm$ 0.29          & \textbf{77,21 $\pm$ 0.09} \\ \hline
MEDICINE    & 58.14 $\pm$ 2.47 & 63.22 $\pm$ 0.47          & \textbf{67,34 $\pm$ 0.51} \\ \hline
MED.PRO.    & 58.51 $\pm$ 0.41 & 59.01 $\pm$ 0.78 & \textbf{61,73 $\pm$ 0.08}          \\ \hline
Mic.F1 & 70.78 $\pm$ 0.20  & 71.22 $\pm$ 0.30          & \textbf{74,78 $\pm$ 0.05} \\ \hline
Mac.F1 & 63.36 $\pm$ 0.70  & 65.28 $\pm$ 0.51          & \textbf{68,76 $\pm$ 0.23} \\ 
\bottomrule[1.5pt]
\end{tabular}
\end{table}
% \subsection{Error Analysis}
% Intent
% In terms of IC, we observe that the majority of mismatch samples are Treatment and Diagnosis intents shows in Table \ref{confusion_matrix}. Due to the ambiguity in Vietnamese, the model still struggles to classify between Treatment or Diagnosis. 
% Besides, the Cause and Severity intents show clear meaning among others. This benefits the model to predict these class easier.
% % table -> figure?

% NER
% With regards to NER, we observe that the number of samples in MEDICINE and MEDICAL$\_$PROCEDURE still shows a limit compare to SYMPTOM\&DISEASE . This may hurt the model due to lacking data.

\section{Discussion}
ViMQ dataset can be utilized to develop a NLU module for healthcare chatbots. We propose a standard use case where it is applicable.
A model trained for NER and IC tasks on the dataset decomposes a question from user to the system into named entities and intent components. These components are used to retrieve a corresponding medical answer from a pre-installed database.
If no respond can be retrieved, the system routes the question to a recommended doctor.

\section{Conclusion}
In this work, we published a Vietnamese dataset of medical questions for the two tasks of intent classification and named entity recognition. Additionally, we proposed a training strategy to learn the model with span-noise modelling. The training strategy demonstrates positive gains on our dataset and COVID-19 Vietnamese NER dataset. Our dataset can be leveraged to develop a NLU module for healthcare chatbots to reduce workload of Telehealth doctors.

% \begin{figure}[!t]
%     \centering
%     \includegraphics[width=\textwidth]{figures/use_case.pdf}
%     \vspace{-0.5cm}
%     \caption{Standard usecase of the ViMQ dataset.}
%     \label{use_case}
% \end{figure}

\bibliographystyle{splncs04}

\bibliography{citation}

%\end{thebibliography}

\end{document}